# A Bayesian Method Reexamined

Derek D. Ayers
rayers@lira.stanford.edu
Department of Engineering-Economic Systems
Stanford University
Stanford, CA 94305-4025

### Abstract

This paper examines the "K2" network scoring metric of Cooper and Herskovits. It shows counterintuitive results from applying this metric to simple networks. One family of noninformative priors is suggested for assigning equal scores to equivalent networks.

## 1 INTRODUCTION

This paper examines the probabilistic network-scoring metric of Cooper and Herskovits [Cooper and Herskovits, 1992; Cooper and Herskovits, 1991; Herskovits, 1991]. This measure is used to distinguish among alternative probabilistic networks, given a database of cases, and is at the heart of a network-construction algorithm called K2. The *K2 metric* emerged from a Bayesian reformulation of an earlier network-construction algorithm called Kutató [Herskovits, 1991; Herskovits and Cooper, 1990], which used an information-theoretic entropy measure. Relationships between the algorithms and their respective metrics are given by Herskovits [Herskovits, 1991].

In addition to a metric for scoring networks, a search is required for selecting candidate networks to evaluate. Network construction proceeds by searching a space of competing network structures and presenting a subset of the networks (typically just one) that score highest. The search is usually greatly constrained by imposing an order on the nodes, and equal prior probabilities on structures commonly are assumed.

The K2 algorithm employs a heuristic myopic-search method that further limits the structure space. K2 starts each node without parents and incrementally adds the parent that gives the highest probability for the resultant structure, until no probability increase occurs from adding any allowable parent. This search also can be truncated when it reaches a predesignated upper bound on the number of parents a node may have.

Different search methods have been proposed, such as applying K2 to several random node orders, or a myopic search for removing parents starting from a maximally connected belief-network structure [Cooper and Herskovits, 1992, page 323]. Others have employed an approach using conditional independence tests to generate a node order, which is then used by the K2 algorithm in constructing a network structure [Singh and Valtorta, 1993].

Though the K2 algorithm and its kin for constructing probabilistic networks are admittedly heuristic in regard to the search problem, the K2 metric is largely accepted for scoring network structures, and the behavior described here is not widely known. The remainder of this paper reviews the K2 metric and presents some examples that show counterintuitive results from applying this metric to simple networks.

## 2 THE K2 METRIC

The probability metric derived by Cooper and Herskovits scores belief-network structures, thereby distinguishing among alternative networks, given a database of cases. Their result is summarized as theorem 1, and is based on the four assumptions that follow.

1. The database variables are discrete.
2. Cases occur independently, given a belief-network model.
3. There are no cases that have variables with missing values.
4. Before observing the database, we are indifferent regarding which numerical probabilities to assign to the belief-network structure.

**Theorem 1** [Cooper and Herskovits, 1992].
Let $Z$ be a set of $n$ discrete variables, where a variable $x_i$ in $Z$ has $r_i$ possible value assignments: $(v_{i1}, \ldots, v_{ir_i})$.
Let $D$ be a database of $m$ cases, where each case contains a value assignment for each variable in $Z$. Let $B_S$ denote a belief-network structure containing just the variables in $Z$. Each variable $x_i$ in $B_S$ has a set of parents, which we represent with a list of variables $\pi_i$. Let $w_{ij}$ denote the $j$th unique instantiation of $\pi_i$ relative to $D$. Suppose there are $q_i$ such unique instantiations of $\pi_i$. Define $N_{ijk}$ to be the number of cases in $D$ in which variable $x_i$ has the value $v_{ik}$ and $\pi_i$ is instantiated as $w_{ij}$. Let $N_{ij} = \sum_{k=1}^{r_i} N_{ijk}$. Given assumptions 1 through 4, it follows that

$$P(B_S, D) = P(B_S) \prod_{i=1}^{n} \prod_{j=1}^{q_i} \frac{(r_i - 1)!}{(N_{ij} + r_i - 1)!} \prod_{k=1}^{r_i} N_{ijk}!.$$
\#

Since the metric is in the form of a joint probability $P(B_S, D)$, the posterior probability $P(B_S | D)$ is proportional to it, so that structures can be ranked as a result. The relative posterior probabilities for two structures can be found by taking the ratio of their respective joint probabilities, since the proportionality constant $P(D)$ will cancel. That is, $P(B_{S_i} | D) / P(B_{S_j} | D) = P(B_{S_i}, D) / P(B_{S_j}, D)$, where $B_{S_i}$ and $B_{S_j}$ are two belief-network structures containing the same set of variables given by the database. The absolute posterior probabilities come at a much greater cost, since obtaining $P(D)$ requires summing $P(B_S, D)$ over all belief-network structures corresponding to the variables in the current database. This latter calculation is tantamount to an exhaustive search; hence, the motivation for constraining the structure space and using heuristics to search for structures having high posterior probabilities.

## 3   EXAMPLES AND DISCUSSION

Ideally, the problem of generating belief-networks from databases can be seen as selecting network structures based on their respective posterior probabilities. If a single most likely structure is desired, this problem can be seen as $B_{S\max} = \mathrm{argmax}_{B_S} [P(B_S | D)]$. Since this involves calculations on the order of exhaustively enumerating all possible structures, it becomes computationally prohibitive even for networks having relatively few nodes. (The number of possible belief-network structures grows at least exponentially with the number of nodes.) However, for extremely small sets of variables, exhaustively enumerating the structures and computing posterior probabilities is quite feasible. The simplest interesting example occurs with two variables, admitting only three belief-network structures. If the two variables are $x_1$ and $x_2$, then the three possibilities are (1) $B_{S_1}$: $x_1$ is the parent of $x_2$, (2) $B_{S_2}$: $x_2$ is the parent of $x_1$, and (3) $B_{S_3}$: neither $x_1$ nor $x_2$ has the other as a parent ($x_1$ and $x_2$ are mutually independent). Using the notation of Cooper and Herskovits [Cooper and Herskovits, 1992, pages 328-329], these can be written as $\{x_1 \to x_2\}$, $\{x_1 \leftarrow x_2\}$, and $\{x_1 \; x_2\}$, respectively.

### 3.1   AN INITIAL EXAMPLE

Table 1: A Nine-Case Database

| variable | |
|---|---|
| $x_1$ | $x_2$ |
| 0 | 0 |
| 1 | 1 |
| -1 | 1 |
| 2 | 2 |
| -2 | 2 |
| 3 | 3 |
| -3 | 3 |
| 4 | 4 |
| -4 | 4 |

Consider the nine-case database represented in table 1 and assume equal prior probabilities over the structures ($P(B_{S_1}) = P(B_{S_2}) = P(B_{S_3}) = 1/3$). Applying theorem 1 to these data, with $r_1 = 9$ and $r_2 = 5$, yields the joint probabilities for each of the three structures: $P(B_{S_1}, D) = 1.935 \times 10^{-17}$, $P(B_{S_2}, D) = 3.481 \times 10^{-17}$, and $P(B_{S_3}, D) = 2.330 \times 10^{-18}$. Summing these over the three belief-network structures gives $P(D) = 5.649 \times 10^{-17}$. Each of the posterior probabilities is obtained by dividing the corresponding joint probability $P(B_S, D)$ by $P(D)$. Thus, $P(B_{S_1} | D) = 0.3425$, $P(B_{S_2} | D) = 0.6163$, and $P(B_{S_3} | D) = 0.0413$.

In this example, $\{x_1 \leftarrow x_2\}$ (i.e., $B_{S_2}$) is scored by the K2 metric as almost twice as likely as $\{x_1 \to x_2\}$ (i.e., $B_{S_1}$), and nearly fifteen times as likely as $\{x_1 \; x_2\}$ (i.e., $B_{S_3}$), for being the belief-network structure that generated the cases in the database. It is not surprising to see that $\{x_1 \; x_2\}$ is unlikely, since there appears to be a rather strong relation between the values of the two variables in the database. An explanation for why $\{x_1 \leftarrow x_2\}$ scores appreciably higher than $\{x_1 \to x_2\}$ is less obvious. (Would *you* want to bet on which structure produced the data, using the odds given by the K2 metric?) An initial reaction is to recognize a functional form between the variables in the database (specifically, $x_2 = |x_1|$), and think of $\{x_1 \to x_2\}$ as the model of the process that generated the data. Indeed, that is precisely how these data were produced. It should be noted that, although the numbers in table 1 are suggestive of absolute value, the exact functional form is not important. For instance, $x_2 = x_1^2$ would give the same results in terms of the K2 metric for the same range of $x_1$, since it is the mapping of categories between variables that is crucial, rather than numerical values. It also can be noted that the relationship between the variables in the database appears as a deterministic one for $\{x_1 \to x_2\}$, but not for $\{x_1 \leftarrow x_2\}$. This is interesting because the K2 metric can be viewed as having a bias against determinism in this instance.

## 3.2 A MORE GENERAL EXAMPLE

Table 2: Generalized Database

| variable | | case counts |
|---|---|---|
| $x_1$ | $x_2$ | ($m$ total) |
| 0 | 0 | $a_0$ |
| 1 | 1 | $a_{1+}$ |
| -1 | 1 | $a_{1-}$ |
| 2 | 2 | $a_{2+}$ |
| -2 | 2 | $a_{2-}$ |
| 3 | 3 | $a_{3+}$ |
| -3 | 3 | $a_{3-}$ |
| 4 | 4 | $a_{4+}$ |
| -4 | 4 | $a_{4-}$ |
| . | . | . |
| . | . | . |
| . | . | . |
| $\omega$ | $\omega$ | $a_{\omega+}$ |
| $-\omega$ | $\omega$ | $a_{\omega-}$ |

The two-variable database of table 1 is generalized in table 2, and is parameterized by an integer vector $A \geq 0$ and an integer $\omega > 0$. $A = (a_0, a_{1+}, a_{1-}, a_{2+}, a_{2-}, \ldots, a_{\omega+}, a_{\omega-})$ is a vector of case counts totaling $m$, where the elements indicate the respective number of occurrences of each pair of values taken on by the variables. For example, $a_0$ is the number of occurrences of $(x_1 = 0, x_2 = 0)$ in the database and $a_{2-}$ is the count of $(x_1 = -2, x_2 = 2)$. The parameter $\omega$ determines the size of the set of possible value assignments for the variables corresponding to the database. For instance, in the initial example, $\omega = 4$, with $a_0 = a_{1+} = a_{1-} = a_{2+} = a_{2-} = a_{3+} = a_{3-} = a_{4+} = a_{4-} = 1$ (call this $A_1$). In terms of theorem 1, $\omega$ implies that $r_1 = 2\omega + 1$ and $r_2 = \omega + 1$.

The initial example can be extended by holding $\omega = 4$ constant while varying $A$. Let $A_{10} = 10 \times A_1$ (i.e., ten occurrences of each case shown in table 1, for a total of $m = 90$ cases in the database). Then $P(B_{S_1} | D) = 0.0727$, $P(B_{S_2} | D) = 0.9273$, and $P(B_{S_3} | D) = 0.0000$ (to four decimal places). Similarly, for $A_{100} = 100 \times A_1$, $P(B_{S_1} | D) = 0.0305$ and $P(B_{S_2} | D) = 0.9695$. For $A_{1000} = 1000 \times A_1$, $P(B_{S_1} | D) = 0.0269$ and $P(B_{S_2} | D) = 0.9731$. Note that the counterintuitive behavior of the K2 metric scoring $\{x_1 \leftarrow x_2\}$ higher than $\{x_1 \rightarrow x_2\}$ becomes even more marked as the number of cases in the database increases. However, even though $P(B_{S_2} | D)$ continues to increase with the size of the database, it does so at a decreasing rate, converging asymptotically to a probability less than 1. For $A_\infty$, the limiting case as the database grows in the same manner without bounds, $P(B_{S_1} | D) = 0.0265$ and $P(B_{S_2} | D) = 0.9735$ (with $P(B_{S_3} | D) \rightarrow 0$; grinding through the factorials gives an exact value, $P(B_{S_2} | D) = 9\,845\,600\,625 / 10\,114\,036\,081$).

Results are more striking when $\omega$ is varied instead of just $A$. Table 3 shows posterior probabilities of $B_{S_1}$, $B_{S_2}$, and $B_{S_3}$ given a variety of databases. The column labeled $A(\omega)$ makes explicit that the length of $A$ is a function of $\omega$, so that $A_1$ indicates that there is a single occurrence of each of the $2\omega + 1$ value-pairs in the database. The main feature illustrated here is that as $\omega \rightarrow \infty$, $P(B_{S_2} | D) \rightarrow 1$ (with $P(B_{S_1} | D) \rightarrow 0$ and $P(B_{S_3} | D) \rightarrow 0$). That is, in the limit as the number of possible value assignments of the variables increases, the K2 metric will score $\{x_1 \rightarrow x_2\}$ with probability zero, even though it is the network structure that actually produced the database! This result holds for finite $A$, but the convergence is even more dramatic when the case counts also grow larger, since the effects of both parameters reinforce one another.

Table 3: Posterior Probabilities After Observing Database

| database parameters | | posterior probabilities | | |
|---|---|---|---|---|
| $\omega$ | $A(\omega)$ | $P(B_{S_1} | D)$ | $P(B_{S_2} | D)$ | $P(B_{S_3} | D)$ |
| 1 | $A_1$ | 0.3600 | 0.4000 | 0.2400 |
| 1 | $A_{10}$ | 0.4172 | 0.5828 | 0.0000 |
| 1 | $A_{100}$ | 0.4020 | 0.5980 | 0.0000 |
| 2 | $A_1$ | 0.3729 | 0.4833 | 0.1438 |
| 2 | $A_{10}$ | 0.2745 | 0.7255 | 0.0000 |
| 2 | $A_{100}$ | 0.2276 | 0.7724 | 0.0000 |
| 4 | $A_1$ | 0.3425 | 0.6163 | 0.0413 |
| 4 | $A_{10}$ | 0.0727 | 0.9273 | 0.0000 |
| 4 | $A_{100}$ | 0.0305 | 0.9695 | 0.0000 |
| 8 | $A_1$ | 0.2205 | 0.7771 | 0.0024 |
| 8 | $A_{10}$ | 0.0023 | 0.9977 | 0.0000 |
| 8 | $A_{100}$ | 0.0001 | 0.9999 | 0.0000 |
| 1c | $A_1$ | 0.0682 | 0.9318 | 0.0000 |
| 16 | $A_{10}$ | 0.0000 | 1.0000 | 0.0000 |

## 4 ANALYSIS

A brief examination of the assumptions of theorem 1 indicates where to look for the source of the observed behavior of the K2 metric. Assumption 1 is that the database variables are discrete. Working only with discrete variables is fairly standard practice, and the symmetry involved (i.e., discrete versus discrete) gives no reason for the observed behavior. Assumption 2 maintains that cases occur independently, given a belief-network model. This expression of conditional independence basically can be interpreted as saying that the database is generated from a stable process (i.e., the

26   Ayersmodel is not changing on us as we receive additional cases), which again does not explain the observed results. The third assumption states that there are no cases that have variables with missing values. This assumption facilitates derivation and computation tasks, but the universality of the database observations makes the assumption irrelevant to the scoring bias exhibited earlier in this paper. That leaves the fourth assumption, which maintains that the density on numerical probabilities is uniform for a given belief-network structure. Assumption 4 is not easy to dismiss, and deserves closer scrutiny.

Cooper and Herskovits leave room for generalizing their scoring metric, call it the *extended K2 metric*, by replacing the uniform density in assumption 4 with a Dirichlet (assumption 4a), which they summarize with the following corollary.

**Corollary 1** [Cooper and Herskovits, 1992].
If assumptions 1, 2, 3, and 4a of theorem 1 hold and second-order probabilities are represented using Dirichlet distributions and $N'_{ij} = \sum_{k=1}^{r_i} N'_{ijk}$, then

$P(B_S, D) =$

$P(B_S) \prod_{i=1}^{n} \prod_{j=1}^{q_i} \frac{(N'_{ij} + r_i - 1)!}{(N_{ij} + N'_{ij} + r_i - 1)!} \prod_{k=1}^{r_i} \frac{(N_{ijk} + N'_{ijk})!}{N'_{ijk}!}$ .

#

The Dirichlet reduces to the uniform in the special case when $N'_{ijk} = 0$ for all valid $i$, $j$, and $k$, and thus corollary 1 reduces to theorem 1 in that circumstance.

In a closely related paper [Buntine, 1991], the form of a prior is proposed which "gives equivalent networks equivalent priors," (presumably he meant that such a prior would give equivalent networks equivalent scores). Similar conditions are stated here which apply directly to the extended K2 metric (that is, when satisfied, they result in the same scores being assigned to equivalent structures):

$$N'_{ijk} = \frac{\alpha}{q_i r_i} - 1 \qquad (*)$$

and

$$N'_{ij} = \frac{\alpha}{q_i} - r_i,$$

for $i = 1, \ldots, n$; $j = 1, \ldots, q_i$; and $k = 1, \ldots, r_i$ (as applicable), and where $\alpha > 0$ has the same value for all $i$, $j$, and $k$. The parameter $\alpha$ controls how strongly to weight prior versus new evidence (*i.e.*, a database). If $\alpha$ is small, the prior is weighted less strongly and the metric is more sensitive to new evidence. If $\alpha$ is large, the prior is weighted more strongly and the metric is less sensitive to new evidence.

If corollary 1 is taken literally, then $N'_{ijk}$ are required to be nonnegative integers, and therefore an additional restriction would be that $\alpha$ is a multiple of $q_i r_i$ for all $i = 1, \ldots, n$. However, it is reasonable to take corollary 1 less literally by replacing the factorials with Gamma functions. Thus, the *generalized K2 metric* can be written as

$P(B_S, D) =$

$P(B_S) \prod_{i=1}^{n} \prod_{j=1}^{q_i} \frac{\Gamma(N'_{ij} + r_i)}{\Gamma(N_{ij} + N'_{ij} + r_i)} \prod_{k=1}^{r_i} \frac{\Gamma(N_{ijk} + N'_{ijk} + 1)}{\Gamma(N'_{ijk} + 1)}$

where the Gamma function is

$$\Gamma(z) = \int_0^\infty t^{z-1} e^{-t} \, dt, \ z > 0.$$

Note that if $z$ is a positive integer, the Gamma function $\Gamma$ has the property that $\Gamma(z) = (z - 1)!$ so it is easy to verify that the generalized K2 metric reduces to the extended K2 metric when all $N'_{ijk}$ are nonnegative integers.

The generalized K2 metric is much less restrictive, allowing the $N'_{ijk}$ to be assigned real values greater than -1. (For $N'_{ijk} = -1$ the factor $\Gamma(N'_{ijk} + 1)$ is not properly defined, though one might choose to think of $N'_{ijk} = -1$ as the limiting extreme of prior ignorance.) This is consistent with the condition $\alpha > 0$, so the conditions (*) are sufficient for giving equivalent networks equal scores with this generalized metric. Table 4 illustrates the generalized K2 metric constrained to give equivalent networks equal scores, using the database from table 1.

Table 4: Equivalent Networks Equally Scored
(Database $D = A_1(4)$ )

| $\alpha$ | $P(B_{S_1} \mid D)$ | $P(B_{S_2} \mid D)$ | $P(B_{S_3} \mid D)$ |
|---|---|---|---|
| 45 | 0.3680 | 0.3680 | 0.2639 |
| 15 | 0.4150 | 0.4150 | 0.1700 |

Notice that these constraints for $N'_{ijk}$ and $N'_{ij}$ do not depend on $j$ and $k$. That is, for each variable $i$, the specific values of the variable and its parents are irrelevant. Instead, the number of possible values that can be taken by the variable (*i.e.*, $r_i$) and its parents (*i.e.*, $q_i$) determine $N'_{ijk}$ and $N'_{ij}$ (along with $\alpha$). In particular, the product $q_i r_i$ can be related to $N'_{ijk}$ in the alternative form

$$N'_{ijk} + 1 \propto \frac{1}{q_i r_i},$$

where $\alpha$ is the same proportionality constant in all instances. Here, $q_i r_i$ is the number of conditional probabilities for variable $i$ with its parents (which is constant, given $i$), so the respective conditional prior probabilities must be equal. Similarly, the corresponding joint probabilities will equal $\frac{1}{q_i r_i}$. Thus, for each variable in the network, the priors are "noninformative" [Berger, 1985, page 82], since no value is favored over others.

The symmetric nature of these noninformative priors seems much in the spirit of the uniform distribution of the original K2 metric. However, in many situations it would be desirable also to use informative priors. Such informative priors could employ constraints which generalize those above in (*).

## 5 CONCLUSION

The special case of the network scoring metric of Cooper and Herskovits that is employed in the K2 search algorithm shows counterintuitive results when applied to simple (two-variable) networks. The behavior described is not completely unknown, but is not widely appreciated. Though the databases presented were generated from a deterministic functional relation, the K2 metric consistently scores against the generating structure in favor of the structure with the arc reversed (which is not a deterministic relation in that form). In the extreme, when the number of possible value assignments of the variables gets large, the generating structure's posterior probability goes to 0. (In this case the variables are becoming closer approximations to continuous variables.) It seems that assigning probability (close to) 0 to the network that produced the database is an undesirable property, and this property is amplified as the number of cases in the database grows. This property is not solely the result of the functional relationships among the variables, but is in large part due to the varying numbers of conditional probabilities between variables and their parents and the corresponding priors chosen for the numerical probabilities in the networks. The observed behavior can be modified through choosing alternative priors in the extended K2 metric, which permits Dirichlet priors in place of the uniform distribution of the basic K2 metric. The metric is also generalized naturally by replacing factorials with Gamma functions appropriately, thereby permitting greater range an flexibility in specifying such priors. Priors should be carefully selected if bias like that which has been observed is to be avoided. A family of noninformative priors is presented that results in equal scores for equivalent networks. Although search methods such as K2 that require specifying a node order in advance may not suffer much in locating relatively high-probability structures, the relative scores should be considered quite dependent on the choice of priors. One should be especially wary of methods which evaluate and compare structures from multiple node orderings, since they are particularly susceptible to the behaviors illustrated in this paper. The results of this paper suggest that researchers may not want to direct effort solely toward searching for high-probability network structures, but in improving the scoring methods used in evaluating such structures and carefully choosing priors.

## Acknowledgments

This research benefited from discussions with colleagues in the Department of Engineering-Economic Systems and from comments of the unnamed reviewers.